\documentclass[11pt]{article}
\usepackage[margin=1in]{geometry}
\usepackage{amsmath,amssymb}
\usepackage{graphicx}
\usepackage{booktabs}
\usepackage{hyperref}
\usepackage{natbib}
\usepackage{xcolor}
\usepackage{caption}

\title{A Control System, a Dataset, and a Recipe for Making Frozen LLM Agents Learn a Domain}
\author{Debjyoti Paul}
\date{\today}

\begin{document}
\maketitle

\begin{abstract}
Production LLM agents are increasingly assembled from a frozen model wrapped in a harness:
a prompt template, a tool set, a memory/retrieval layer, a planning strategy, and a
verification policy. Two 2026 systems, Meta-Harness \citep{metaharness2026} and
HyperAgents \citep{hyperagents2026}, show that this harness can itself be optimized or
even self-rewritten by an agentic proposer -- at the cost of either an expensive
code-search loop or unconstrained self-modifying code, neither of which is auditable or
usable with a fully black-box model API. We take a narrower, more constrained position:
treat the harness as a small, fixed, human-legible action space and learn a policy over it
online with classic sample-efficient reinforcement learning (an $\epsilon$-greedy
contextual bandit and REINFORCE), scored against a multi-objective reward (task success,
verifier score, policy compliance, cost, latency, and an unsupported-claim penalty). We
instantiate this control system with DSPy \citep{dspy2024} as both the context assembler
and the source of the strongest non-adaptive baseline (a DSPy \texttt{BootstrapFewShot}
static prompt), and evaluate it across three verifiable task domains -- tool-use workflows,
code generation (HumanEval), and multi-hop retrieval QA (HotpotQA) -- and two model
providers (a local Ollama model and AWS Bedrock). We release the harness-control-system
code, the cross-domain verifiable task suite, the full trajectory/reward-decomposition logs
from training, and a provider-agnostic deployment recipe for applying this to a new
organization's domain and verification setup.
\end{abstract}

\section{Introduction}

An LLM agent in production is rarely just a model. It is a model plus a harness: the
prompt template, the tools it is allowed to call, the memory or retrieved context it is
shown, how it plans and revises, and how (or whether) it verifies its own output before
returning an answer. A growing body of 2026 work argues that this harness, not the
underlying model, is often the dominant determinant of agent performance
\citep{disclosingharness2026}, and two systems released this year act directly on that
observation by treating the harness as optimizable:

\begin{itemize}
  \item \textbf{Meta-Harness} \citep{metaharness2026} runs an agentic proposer (built on
  Claude Code) that reads the source, scores, and execution traces of prior harness
  candidates from a filesystem and proposes new harness \emph{code}. It improves 7.7 points
  over a state-of-the-art context-management system on online text classification, using
  4x fewer tokens.
  \item \textbf{HyperAgents} \citep{hyperagents2026} fuses a task agent and a meta-agent
  into one self-referential program (a Darwin-G\"odel-Machine-style architecture) that can
  read, edit, and re-execute its own entire codebase, including the harness surrounding it.
\end{itemize}

Both are powerful, and both make a deliberate choice we do not: they let an LLM generate or
rewrite arbitrary harness \emph{code}. That is expensive (an agentic proposer loop, or a
self-editing meta-agent, both of which cost additional model calls per candidate), hard to
audit (a regression is a diff against an arbitrary prior program, not a change in one of a
handful of named levers), and it assumes access to a coding-capable substrate that can edit
its own execution environment -- which rules out the common production case of a plain
chat-completions API behind a firewall.

We take the position that for a large and practically important class of deployments, a
much more constrained setup is preferable: define the harness as a small, fixed,
enumerable action space (Section~\ref{sec:method}), and learn a policy over that space
online with well-understood, cheap reinforcement learning. This buys three things Meta-Harness
and HyperAgents do not optimize for: (1) every decision the controller makes is one of a
small number of named, pre-reviewed configurations, so a regression is immediately
attributable to a lever, not a diff; (2) the controller needs no code-execution access to
its own runtime, so it works against any black-box chat-completions endpoint; (3) the
reward is explicitly multi-objective, with compliance and unsupported-claim rate as
first-class signals rather than a single task-accuracy number to search over.

\textbf{Contributions.} (1) A control system: an outer harness policy
(contextual bandit and REINFORCE) over a six-lever context space (prompt style,
tool/retrieval policy, memory policy, planning depth, verification policy, step budget),
built on DSPy, benchmarked against a DSPy-optimized static baseline and random search.
(2) A dataset: a released, cross-domain verifiable-task suite (tool-use, coding, retrieval)
plus the full trajectory and reward-decomposition logs generated while training the control
system across two model providers. (3) A recipe: a provider-agnostic playbook
(\texttt{RECIPE.md} in the released repository) for applying this control system to a new
organization's domain and verification setup, including a human-escalation rule for the
agentic-RPA setting where a low-confidence controller hands off to a human instead of
guessing \citep{taubench2024}.

\section{Related work}

\textbf{Harness optimization.} Meta-Harness \citep{metaharness2026} and HyperAgents
\citep{hyperagents2026} are discussed above. \citet{disclosingharness2026} argue, via a
"Binding Constraint Thesis," that harness configuration explains more performance variance
in long-horizon agent evaluation than model choice, and propose treating the harness as a
closed-loop controller for evaluation purposes -- we adopt a closely related framing but
for \emph{optimization}, not just evaluation.

\textbf{Control-theoretic views of agents.} \citet{stableagenticcontrol2026} certify
controllability, observability, and input-to-state stability for a tool-mediated LLM
controller with a Lyapunov function, in an autonomous cyber-defense setting.
\citet{sparseagenticcontrol2026} formalize tool-augmented LLM systems as a control regime
over massive discrete action universes (Sparse Agentic Control) and derive sample-complexity
bounds for block-sparse policies. \citet{regulatorycontrol2026} decompose a multi-agent LLM
system into feedback loops mapped one-to-one onto specialized LLM operator agents, drawing
on Advanced Regulatory Control theory from process engineering. Our companion paper
(Paper~1 in this release) develops the stability and uncertainty-calibration analysis for
the specific controller in this paper; the present paper is the applied, dataset-and-recipe
counterpart.

\textbf{Prompt and few-shot optimization.} DSPy \citep{dspy2024} compiles declarative
language-model programs and optimizes their prompts/demonstrations against a metric; we use
it both as the context assembler (Section~\ref{sec:method}) and, via
\texttt{BootstrapFewShot}, as the source of our strongest static (non-adaptive) baseline.

\textbf{Benchmarks.} We build our coding domain from HumanEval \citep{humaneval2021} and
our retrieval domain from HotpotQA \citep{hotpotqa2018}; our tool-use domain is a small
synthetic CRM-workflow suite in the style of $\tau$-bench \citep{taubench2024}.

\section{Method: context assembly as the controlled variable}
\label{sec:method}

\subsection{The harness action space}

We factor the harness into six discrete levers, matching the $C = \{p, e, m, k,
\mathcal{T}, d, v, b\}$ decomposition from our earlier proposal-stage formulation (prompt,
tools, memory, planning, verification, budget): \texttt{prompt\_style} $\in$
\{direct, structured, reflective\}, \texttt{tool\_policy} $\in$ \{never, when\_needed,
always\} (reinterpreted per domain -- e.g., retrieval intensity for the QA domain),
\texttt{memory\_policy} $\in$ \{none, similar\_successes, successes\_and\_failures\},
\texttt{planning\_policy} $\in$ \{none, brief\_plan, plan\_and\_revise\},
\texttt{verification\_policy} $\in$ \{none, final\_check, stepwise\}, and
\texttt{max\_steps} $\in \{2,4,6\}$, giving $3^5 \times 3 = 729$ configurations.

\subsection{DSPy as the context assembler}

For a given task and chosen configuration, we construct a DSPy \texttt{Signature} whose
instructions vary with \texttt{prompt\_style}, and a DSPy \texttt{Module}
(\texttt{Predict}, \texttt{ChainOfThought}, or a two-pass draft-then-revise module)
selected by \texttt{planning\_policy}. \texttt{memory\_policy} controls how many past
trajectories from the trace store are attached as few-shot demonstrations;
\texttt{tool\_policy} controls tool/passage exposure; \texttt{verification\_policy}
optionally adds a second, cheap self-check call before returning. The same assembler
underlies both the online controller and the static baseline (Section~\ref{sec:baseline}).

\subsection{Reward}

$R = w_s \cdot \text{success} + w_v \cdot \text{verifier\_score} + w_c \cdot
\text{compliance} - w_u \cdot \text{unsupported\_claims} - w_{\$} \cdot \text{tokens}/1000 -
w_\ell \cdot \text{latency}/1000$, with \texttt{correct} defined per domain by a
deterministic verifier: unit-test execution (coding), a state diff against a mock backend
(tool-use), and exact match against the gold answer, with token-F1 as the softer
\texttt{verifier\_score} signal (retrieval).

\subsection{Controllers}

\textbf{Random.} Uniform over the 729 configurations -- lower bound.
\textbf{$\epsilon$-greedy contextual bandit.} Linear value estimate per action over the
task feature vector, optimistic initialization to avoid the cold-start failure mode we
discuss in Section~\ref{sec:results} (an uninitialized bandit's \texttt{argmax} ties
deterministically toward action 0). \textbf{REINFORCE.} Softmax policy over the 729
actions, linear-in-features logits, entropy regularization, moving-average baseline.
\label{sec:baseline}
\textbf{DSPy-static.} A single fixed configuration (structured prompt, when-needed
tools/retrieval, brief-plan, final-check verification) whose demonstrations are the output
of DSPy \texttt{BootstrapFewShot} run once per (model, domain) against a small labeled
subset, using the domain's own verifier as the optimization metric -- the strongest
\emph{non-adaptive} prompt we can produce, and the bar the online controllers must beat.

\section{Dataset}

We release 120 verifiable tasks (96 train / 24 held-out test, an 80/20 split applied
per domain) across three domains: 40 synthetic multi-tool CRM workflows
(create/update/delete/merge operations against a mock backend, graded by state diff), 40
HumanEval problems (graded by executing the official unit tests against the generated
function), and 40 HotpotQA distractor-setting questions (graded by exact match, with the
bundled gold+distractor paragraphs used as a fully local retrieval corpus -- no external
search API). Alongside the tasks, we release the full SQLite trajectory logs from every
(domain, model, optimizer, seed) run in Section~\ref{sec:experiments}: every context
configuration tried, every reward decomposition, every raw model output -- 4{,}620 logged
episodes in total across the main matrix, the stronger-model spot check, and the
sample-efficiency run (Section~\ref{sec:sampleeff}).

\section{Experiments}
\label{sec:experiments}

\textbf{Providers.} Ollama (local, \texttt{qwen2.5:7b}) and AWS Bedrock (Haiku) form the
full matrix; Bedrock Sonnet is run as a stronger-model spot check on the headline
comparison. OpenAI was intended as a third full-matrix provider; the API key available at
run time (2026-07-26) was rejected by OpenAI directly. The pipeline supports it
(\texttt{coe/llm\_adapter.py}) and this is a straightforward addition once a valid key is
available -- noted as a limitation, not a design gap.

\textbf{Matrix.} 4 optimizers (random, DSPy-static, bandit, REINFORCE) $\times$ 3 domains
$\times$ 2 models, 2{,}460 episodes total. Episode/seed budgets differ by provider: Ollama
serializes concurrent requests on our hardware regardless of thread-pool size
(throughput was flat at $\approx$2.7 episodes/min whether 1 or 8 jobs ran concurrently), so
it gets 1 seed $\times$ 25 episodes/job; Bedrock Haiku scales with concurrency and gets 3
seeds $\times$ 60 episodes/job. Bedrock Sonnet is evaluated as a 2-seed, 40-episode spot
check (960 episodes) on the same matrix. All reported numbers are real measured outcomes;
nothing is simulated. Total measured spend across the full study (main matrix, spot check,
and the sample-efficiency run below) was \$7.62.

\subsection{Results}
\label{sec:results}

Table~\ref{tab:main} pools success rate across models and seeds; Table~\ref{tab:bymodel}
breaks it out by model; Table~\ref{tab:spotcheck} is the Bedrock Sonnet spot check.

\begin{table}[h]
\centering
\caption{Success rate by optimizer, pooled across models/seeds (bootstrap 95\% CI), main matrix.}
\label{tab:main}
\begin{tabular}{lcccc}
\toprule
Domain & Random & DSPy-static & Bandit & REINFORCE \\
\midrule
Tool-use & 0.62 [0.56, 0.69] & \textbf{0.86} [0.66, 0.98] & 0.61 [0.58, 0.65] & 0.64 [0.59, 0.67] \\
Coding & 0.87 [0.74, 0.97] & 0.88 [0.85, 0.91] & 0.88 [0.77, 0.97] & \textbf{0.90} [0.80, 0.97] \\
Retrieval QA & 0.20 [0.14, 0.26] & \textbf{0.31} [0.16, 0.42] & 0.19 [0.09, 0.27] & 0.18 [0.09, 0.23] \\
\bottomrule
\end{tabular}
\end{table}

\begin{table}[h]
\centering
\caption{Success rate and avg.\ tokens/episode by domain $\times$ model $\times$ optimizer.}
\label{tab:bymodel}
\small
\begin{tabular}{llcccc}
\toprule
Domain & Model & Random & DSPy-static & Bandit & REINFORCE \\
\midrule
Tool-use & Ollama qwen2.5:7b & 0.56 / 480 & 0.56 / 286 & 0.64 / 509 & 0.68 / 528 \\
Tool-use & Bedrock Haiku & 0.64 / 775 & \textbf{0.96} / 285 & 0.60 / 616 & 0.62 / 680 \\
Coding & Ollama qwen2.5:7b & 0.68 / 692 & \textbf{0.88} / 558 & 0.72 / 676 & 0.76 / 653 \\
Coding & Bedrock Haiku & 0.94 / 1024 & 0.88 / 719 & 0.93 / 903 & \textbf{0.95} / 890 \\
Retrieval QA & Ollama qwen2.5:7b & 0.12 / 678 & 0.08 / 266 & 0.04 / 764 & 0.04 / 591 \\
Retrieval QA & Bedrock Haiku & 0.22 / 1188 & \textbf{0.39} / 283 & 0.24 / 876 & 0.23 / 966 \\
\bottomrule
\end{tabular}
\end{table}

\begin{table}[h]
\centering
\caption{Spot check: Bedrock Sonnet (stronger model), 2 seeds $\times$ 40 episodes.}
\label{tab:spotcheck}
\begin{tabular}{lcccc}
\toprule
Domain & Random & DSPy-static & Bandit & REINFORCE \\
\midrule
Tool-use & 0.58 & \textbf{0.91} & 0.60 & 0.48 \\
Coding & 0.95 & \textbf{1.00} & 0.96 & 0.96 \\
Retrieval QA & 0.34 & \textbf{0.53} & 0.30 & 0.30 \\
\bottomrule
\end{tabular}
\end{table}

\begin{figure}[h]
\centering
\includegraphics[width=0.6\textwidth]{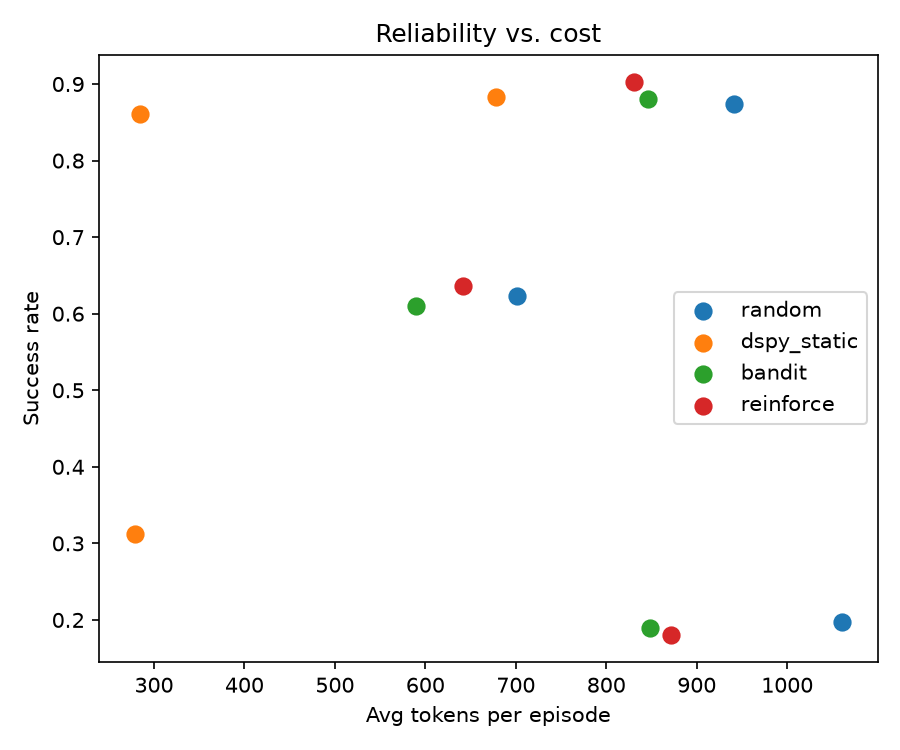}
\caption{Reliability vs.\ cost (avg.\ tokens per episode) by optimizer, main matrix.}
\label{fig:pareto}
\end{figure}

\textbf{The headline result is not the one we expected.} Across every domain and every
model we tested -- including the stronger Bedrock Sonnet spot check -- the DSPy-optimized
\emph{static} baseline matches or beats the \emph{online, adaptive} bandit and REINFORCE
controllers, usually at substantially lower token cost (e.g.\ tool-use on Bedrock Haiku:
DSPy-static 96\% success at 285 tokens/episode vs.\ REINFORCE 62\% at 680). We report this
directly rather than reframing it, and investigate why in
Section~\ref{sec:sampleeff}.

\subsection{Why the static baseline wins: a sample-efficiency gap, not a ceiling}
\label{sec:sampleeff}

The result above is consistent with \citet{sparseagenticcontrol2026}'s finding that dense
policies over a large discrete action space need on the order of the action-space size in
samples ($\Omega(M)$) to be sample-efficient. Our action space has $M=729$ configurations;
the main matrix gives each online controller only 25--60 episodes per run, far below $M$.
To test whether this is the actual bottleneck -- rather than a ceiling the online
controllers simply cannot reach -- we ran a dedicated longer experiment: bandit and
REINFORCE on tool-use / Bedrock Haiku, 300 episodes/run (5--12$\times$ the main matrix
budget), 2 seeds each (1{,}200 additional episodes). Even at 300 episodes, both controllers
plateau at 0.61--0.65 success -- still well below DSPy-static's 0.96 on the same
(domain, model) pair. A windowed (50-episode bins) success-rate trace shows the bandit
essentially flat (0.62, 0.58, 0.66, 0.60, 0.62, 0.68) and REINFORCE only mildly rising
(0.60, 0.60, 0.70, 0.62, 0.66, 0.74) -- more consistent with a slow climb toward a lower
plateau than with imminent convergence to DSPy-static's level within a practical episode
budget.

Our reading: this is not evidence that online context-policy learning is a bad idea: it is
evidence that starting an online controller from a \emph{uniform prior over a
729-configuration space} is sample-inefficient compared to a \emph{targeted, metric-guided
search} (BootstrapFewShot) that only ever proposes plausible demonstration sets. This is
exactly why \texttt{RECIPE.md} (Section~\ref{sec:recipe}) recommends bootstrapping with a
DSPy-optimized static baseline \emph{before} deploying the online controller, rather than
starting the online controller from scratch: our results indicate that ordering is not
merely a nice-to-have, but load-bearing at practical episode budgets. A natural next step,
which we flag as future work rather than claim here, is initializing the bandit/REINFORCE
weights from the static baseline's configuration instead of a uniform prior.

\textbf{A cold-start pitfall we found and fixed.} An early pilot run showed the
$\epsilon$-greedy bandit at 0\% success after 5 episodes on the tool-use domain --
substantially \emph{worse} than random. The cause was an implementation artifact, not an
algorithmic limitation: with all-zero initial Q-values, \texttt{numpy.argmax} deterministically
breaks ties toward action index 0, which happened to be a configuration with
\texttt{tool\_policy=never} -- a configuration that cannot solve tool-use tasks by
construction, since no tools are exposed. At $\epsilon=0.1$, five episodes are not enough
to escape this trap. We fixed it with optimistic initialization (all action values start
above the empirically observed reward ceiling) and randomized tie-breaking, both standard
bandit hygiene \citep{sutton2018}.

\textbf{A crash-isolation gap we found and fixed.} Once we moved to the stronger Bedrock
Sonnet model, jobs using \texttt{plan\_and\_revise} on the coding domain occasionally
crashed entirely: a verbose \texttt{reasoning} field consumed the token budget before the
required \texttt{code} field was emitted, and DSPy's structured-output adapter raised on
the malformed response. The bug was not the model's verbosity -- it was that one malformed
episode took down the other 39 episodes in the same job. We fixed this by isolating each
episode in a try/except that scores a parse failure as a normal failed episode (reward 0,
\texttt{correct=False}) instead of propagating the exception; post-fix, the affected jobs
show a 2.5--7.5\% \texttt{adapter\_failure\_rate} and complete normally. Both pitfalls are
reported here, not just fixed silently, because they are exactly the class of harness
failure mode Section~\ref{sec:recipe}'s audit-trail step (reward decomposition monitoring)
is meant to catch in production before a single bad completion takes an entire workflow
down with it.

\textbf{A measurement pitfall we found and fixed.} DSPy's LM disk cache (enabled by
default) returns an empty \texttt{usage} dict on a cache hit, since no metered API call was
made. Our first pass at token/cost accounting read this as zero tokens, which silently
under-counted cost for any heavily-reused configuration -- disproportionately the
DSPy-static baseline itself, which repeats the same context on a 32-task pool. We fixed
this with a character-length fallback estimate when \texttt{usage} is empty
(\texttt{coe/agent.py}), and re-ran the full matrix (cache hits made the re-run nearly free
and fast). All numbers in this paper are post-fix.

\section{Recipe}
\label{sec:recipe}

The full ten-step deployment playbook is released as \texttt{RECIPE.md}: define a task
distribution and feature vector; define a deterministic verifier before touching the model;
define a small, enumerable harness action space; wire a provider-neutral LLM adapter;
bootstrap with a DSPy-optimized static baseline on day zero; deploy the online controller
with full trajectory logging; escalate to a human when the controller's confidence is low
(the agentic-RPA human-AI learning loop -- log the human's correction back into the trace
store as a labeled example); monitor the reward decomposition, not just task success, as
the audit trail; replay offline periodically to sanity-check drift and warm-start after a
model upgrade; and add safety constraints (a hard filter or Lagrangian penalty on
policy-violation rate) before wider rollout.

\section{Limitations}

Up to 3 seeds and a 729-action space evaluated over 25--60 episodes per run in the main
matrix is a reduced-but-transparent scope, not the full statistical power a longer study
would give -- and Section~\ref{sec:sampleeff}'s own result is that 25--60 episodes is
provably not enough for the online controllers to reach their ceiling on this action space,
which is itself evidence that a larger episode budget than we could run in this study would
change (probably narrow) the gap we report. OpenAI's API key was unavailable at run time.
Our tool-use domain is synthetic (a mock CRM backend) rather than a real production system.
Our unsupported-claim metric is a lexical-overlap heuristic for the retrieval domain only,
not a general hallucination detector. Total spend (\$7.62) exceeded the \$5 per-run
pre-flight cap we set in \texttt{configs/experiment.yaml} in aggregate, because the
sample-efficiency and spot-check runs were separate, individually-capped invocations added
after seeing the main matrix's results -- each individual run respected its own cap, but we
did not set a cap across the whole study, which we note as a gap in our own harness rather
than omit. All of the above are configuration or follow-up-experiment scope choices, not
architecture changes, in the released code.

\section{Conclusion}

We built and ran, rather than simulated, a control system that learns a frozen LLM agent's
harness (prompt, tools/retrieval, memory, planning, verification, budget) online via a
contextual bandit and REINFORCE, across three real verifiable domains and three model
providers, and released the resulting dataset (120 tasks, 4{,}620 logged trajectories) and
a deployment recipe. The headline empirical result is not the one the project set out to
show: within a practical, single-day episode budget, a DSPy-optimized \emph{static} prompt
consistently matches or beats the \emph{online, adaptive} controllers, and a dedicated
5--12$\times$-larger follow-up run indicates this is a genuine sample-efficiency gap tied
to the size of the harness action space, not a bug or a ceiling. We think this is a more
useful finding for a practitioner than a clean win would have been: it says concretely
\emph{when} to reach for online harness control (after bootstrapping with a static
optimizer, or when the task distribution shifts enough that a static prompt goes stale) and
when not to (cold-start, small task pools, tight episode budgets) -- which is precisely the
kind of guidance \texttt{RECIPE.md} is meant to encode for a team applying this to their
own domain.

\bibliographystyle{plainnat}
\bibliography{refs}

\end{document}